\pdfoutput=1

\documentclass[11pt]{article}

\usepackage[preprint]{acl}

\usepackage{times}
\usepackage{latexsym}
\usepackage{tabularx}
\usepackage{makecell}
\usepackage{multirow}
\usepackage{amssymb}
\usepackage{amsmath}
\usepackage{bm}
\usepackage{subfigure}
\usepackage{booktabs} 
\usepackage[T1]{fontenc}

\usepackage[utf8]{inputenc}

\usepackage{microtype}

\usepackage{inconsolata}

\usepackage{graphicx}

\title{Mixture Policy based Multi-Hop Reasoning over N-tuple\\ Temporal Knowledge Graphs}

\author{
\textbf{Zhongni Hou}\textsuperscript{1,2},
\textbf{Miao Su}\textsuperscript{1,2},
\textbf{Xiaolong Jin}\textsuperscript{1,2},
\textbf{Zixuan Li}\textsuperscript{1,2},\\
\textbf{Long Bai}\textsuperscript{1,2},
\textbf{Jiafeng Guo}\textsuperscript{1,2},
\textbf{Xueqi Cheng}\textsuperscript{1,2} \\
\textsuperscript{1}CAS Key Lab of Network Data Science and Technology, ICT, CAS \\
\textsuperscript{2}University of Chinese Academy of Sciences \\
\texttt{zhongni.hou@gmail.com}, \texttt{\{sumiao22z, jinxiaolong, lizixuan, bailong, guojiafeng, cxq\}@ict.ac.cn}
}

\begin{document}
\maketitle
\begin{abstract}
Temporal Knowledge Graphs (TKGs), which utilize quadruples in the form of (subject, predicate, object, timestamp) to describe temporal facts, have attracted extensive attention.
N-tuple TKGs (N-TKGs) further extend traditional TKGs by utilizing n-tuples to incorporate auxiliary elements alongside core elements (i.e., subject, predicate, and object) of facts, so as to represent them in a more fine-grained manner. 
Reasoning over N-TKGs aims to predict potential future facts based on historical ones. However, existing N-TKG reasoning methods often lack explainability due to their black-box nature. 
Therefore, we introduce a new Reinforcement Learning-based method, named MT-Path, which leverages the temporal information to traverse historical n-tuples and construct a temporal reasoning path.
Specifically, in order to integrate the information encapsulated within n-tuples, i.e., the entity-irrelevant information within the predicate, the information about core elements, and the complete information about the entire n-tuples, MT-Path utilizes a mixture policy-driven action selector, which bases on three low-level policies, namely, the predicate-focused policy, the core-element-focused policy and the whole-fact-focused policy. 
Further, MT-Path utilizes an auxiliary element-aware GCN to capture the rich semantic dependencies among facts, thereby enabling the agent to gain a deep understanding of each n-tuple.
Experimental results demonstrate the effectiveness and the explainability of MT-Path.

\end{abstract}

\section{Introduction}
Knowledge Graphs (KGs), which store facts in the form of triples, i.e., (subject, predicate, object), are the key infrastructure of modern AI~\cite{he2017learning,yang2022knowledge}. However, facts usually change over time, and the corresponding validity may also change. To characterize the temporal dynamics of facts, Temporal KGs (TKGs) additionally associate each triplet with a timestamp as a quadruple, i.e., (subject, predicate, object, timestamp)~\cite{li2022hismatch,zhang2023learning}. 

\begin{figure}[t]
  \centering  
   \includegraphics[width=0.8\linewidth]{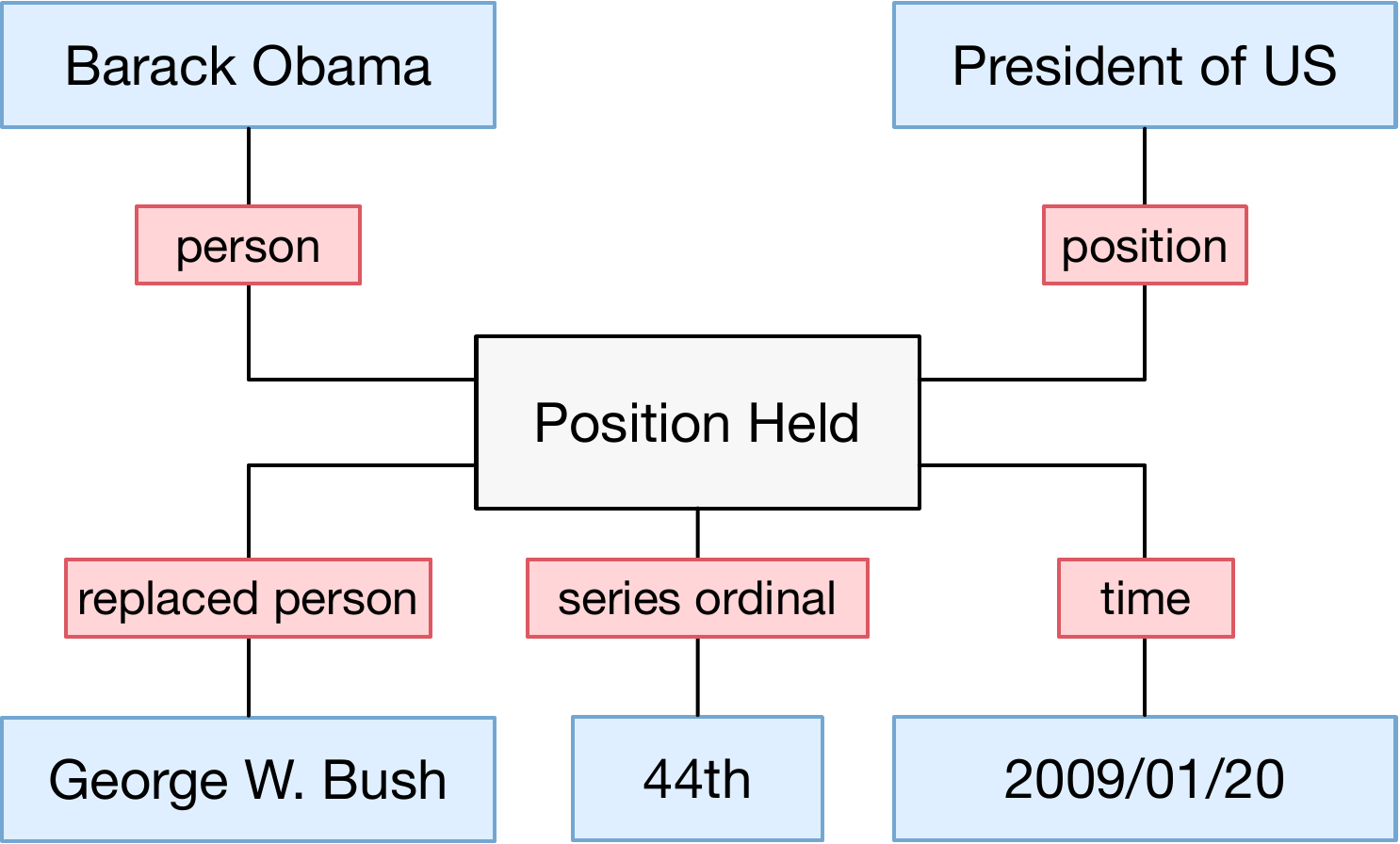}
   \caption{The illustration of an n-tuple fact {Barack Obama replaced George W. Bush as the 44th US president on 20 January 2009}. This fact can be represented as {Position Held (person: Barack Obama, position: President of US, replaces: George W. Bush, series ordinal: 44th, 2009/01/20)}.}
   \label{fig:intro}
    \vspace{-7mm}
  \end{figure}
 
Despite its wide adoption, the quadruple-based representation has no means for representing additional information of facts~\cite{hou2023temporal,ding2023exploring}. Actually, except for the core elements (i.e., subject, predicate, and object), facts usually involve other auxiliary arguments (i.e., entities) and their corresponding roles. As illustrated in Figure~\ref{fig:intro}, {Barack Obama} held the position as {president} of the {US} on {2009/01/20}. In addition, this fact also involves two additional entities, i.e., \emph{44th} and \emph{George W. Bush}, playing the corresponding roles, i.e., the series ordinal and the replaced person, respectively.  To overcome the shortcoming of the quadruple-based representation, N-tuple TKGs (N-TKGs) have been proposed, wherein each fact is represented as an n-tuple, i.e., predicate (role$_1$:entity$_1$, ..., timestamp)~\cite{ding2023exploring,hou2023temporal}. Reasoning over N-TKGs that aims to predict future potential facts based on historical ones, such as \emph{Coop (Cooper1: America, Cooper2: ?, Content: Economic, 2026-10-01)}. Obviously, such a task may be helpful for many time-sensitive applications, such as policymaking~\cite{deng2020dynamic} and disaster relief~\cite{signorini2011use}.

To conduct this task, several N-TKG reasoning methods~\cite{hou2023temporal,ding2023exploring} have been recently proposed. Unfortunately, these methods are inherently embedding-based and operate as black-box models, lacking the ability to provide interpretation for a given prediction. An alternative solution for N-TKG reasoning is to make predictions via synthesizing information from multi-hop temporal paths containing n-tuples, e.g., \emph{Express Intent to Coop (Cooper1: Japan, Cooper2: America, content: Military, $t-1$) $\wedge$ Consult (Consulter: America, Consulted: Japan, Consult way: Visit, $t-2$)}. However, how to conduct the multi-hop reasoning over N-TKGs has not been explored. 

The primary challenge lies in how to effectively utilize the time information and the diverse semantic information within n-tuples. 
Typically, more recent facts carry greater significance than those from the distant past. Regarding the semantic information, the core elements provide essential information about an n-tuple, whereas the auxiliary elements offer supplementary descriptive details~\cite{guan2020neuinfer,rosso2020beyond}. 
It is notable that the contribution of auxiliary elements to correct predictions can vary significantly across different queries. For instance, over-focusing on the location within a \emph{Make a Statement} fact can be useless. 
Additionally, new entities may emerge over time due to the dynamic nature of N-TKGs. When predicting facts involving these unseen entities, relying solely on the information within these entities is challenging, since the model lacks prior knowledge about them. 

With these considerations in mind, in this paper, we propose a Reinforcement Learning (RL)-based method, called Mixture-policy Time-aware Path (MT-Path), to mine temporal patterns and conduct multi-hop reasoning over historical n-tuples. 
Specifically, MT-Path focuses on the action selection strategy and leverages a mixture policy-driven action selector to determine the next hop. 
At each step, MT-Path decomposes the action selection strategy into three low-level ones, i.e., the predicate-focused policy, the core-element-focused policy, and the whole-fact-focused policy, to capture both the time information as well as the diverse semantics in n-tuples, i.e., the entity-irrelevant information within the predicate, the information about core elements, and the complete information about the entire n-tuples, respectively. The former policy allows the model to make precise predictions when facing queries containing unseen entities. The latter two policies highlight the semantic information within the core elements and simultaneously incorporate the complete information of a fact.
Subsequently, an MLP-based gate is utilized to adaptively aggregate the action score calculated by each low-level policy. 
Considering that there are rich semantic dependencies among historical facts, which can provide more comprehensive information for facts, MT-Path further introduces an auxiliary element-aware GCN to model these complex dependencies.

In summary, our main contributions are as follows: (1) We first utilize the RL framework for reasoning over historical n-tuples, which can provide explainable paths for different predictions; (2) We propose a RL-based method, named MT-Path, which employs a mixture policy-driven action selector and an auxiliary element-aware GCN to conduct reasoning over N-TKGs. The former decomposes the action selection strategy into three low-level ones to synthesize the entity-irrelevant information within the predicate, the information about the core elements, and the complete information about the entire fact. The latter captures the semantic dependencies within historical facts, which enables the agent to understand each fact more thoroughly; (3) Experimental results show the effectiveness and the explainability of MT-Path on N-TKG reasoning.

\section{Related Work}
{\bf Static N-tuple Reasoning.}
Static n-tuple KG reasoning aims to infer the missing elements of a given n-tuple. 
Based on the multiple role-entity pairs in n-tuples, some recent works~\cite{guan2019link,liu2021role} try to learn the relatedness between the role and the entity to conduct reasoning. 
Among them, NaLP~\cite{guan2019link} and tNaLP~\cite{guan2021link} measure the plausibility of each n-tuple fact via modeling the compatibility between the role and different entities through a CNN. 
RAM~\cite{liu2021role} further enforces semantically related roles to share similar representations, and employs the inner product to model the relatedness between a role and all involved values.
On the other hand, recent studies~\cite{rosso2020beyond,wang2021link,hu2023hyperformer,shomer2023learning} point out that each element in n-tuples is of different importance, and propose to represent an n-tuple as the combination of a main triplet and auxiliary role-entity pairs. 
For example, HINGE~\cite{rosso2020beyond} and NeuInfer~\cite{guan2020neuinfer} design two different feature extraction pipelines for the main triplet and role-entity pairs, respectively.
StarE~\cite{galkin2020message} leverages CompGCN~\cite{vashishth2019composition} to encode the information within the auxiliary role-entity pairs, and subsequently integrates it with predicate and the entity embeddings to perform reasoning. 
However, none of these models focus on temporal reasoning, and thus cannot appropriately handle the N-TKG reasoning task.

\begin{figure*}[t]
  \centering  
   \includegraphics[width=0.85\linewidth]{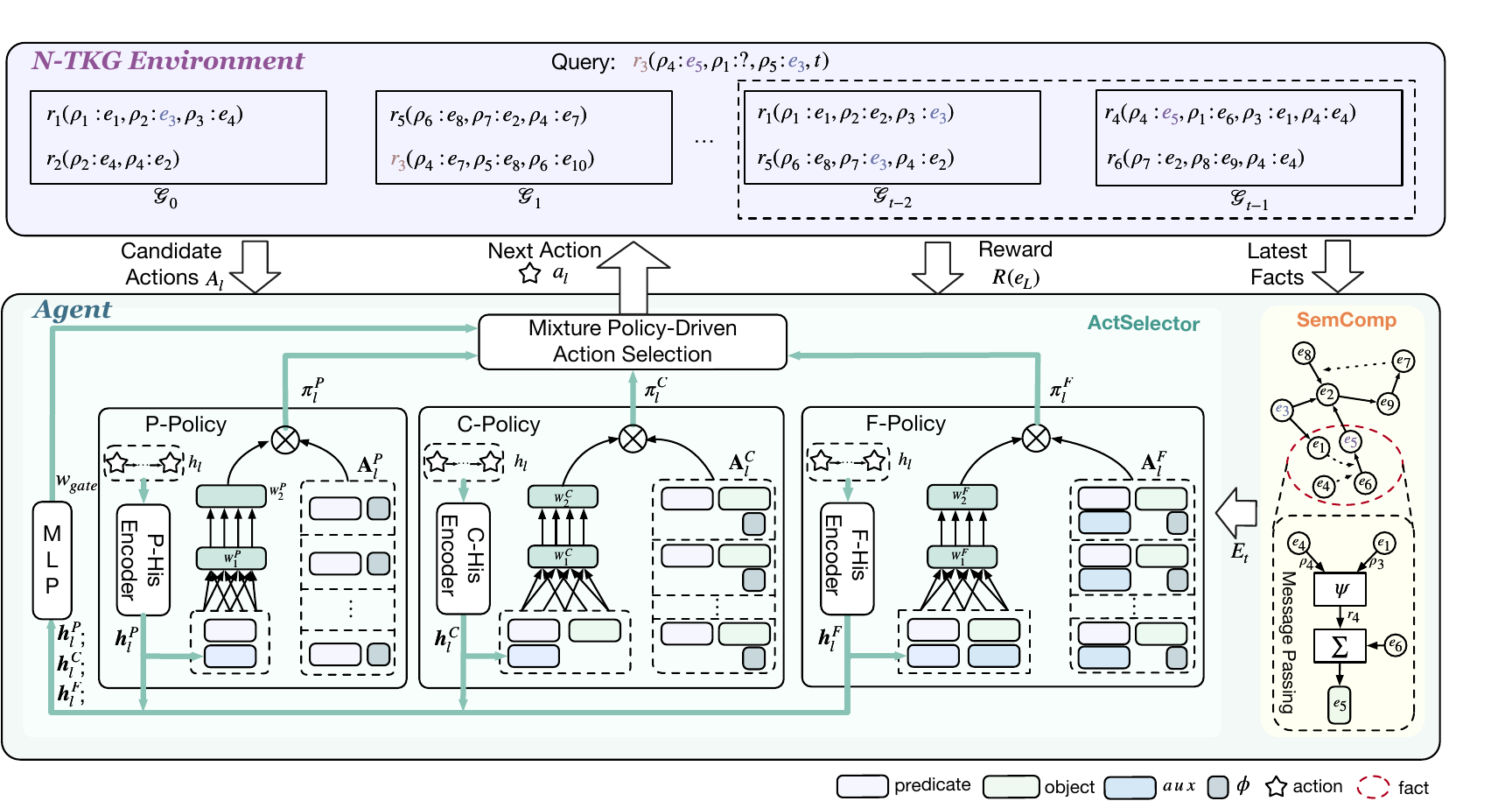}
   \caption{An illustrative diagram of the proposed MT-Path model for multi-hop reasoning over N-TKGs. For the sake of brevity, each historical n-tuple's occurring time is not explicitly given.}
   \label{fig:model}
   \vspace{-5mm}
  \end{figure*}

{\bf TKG Reasoning and N-TKG Reasoning.}
TKG reasoning aims to predict further facts by leveraging historical quadruples, whereas N-TKG reasoning endeavors to make predictions based on historical n-tuples. 
A prevailing approach for TKG reasoning is incorporating TKG embedding~\cite{han2020explainable,zhu2020learning,han2021learning,park2022evokg,li2022tirgn,chen2021dacha,liang2023learn,tang2023gtrl,tang2024dhyper,liu2024temporal}, which learns embeddings for entities and predicates across different timestamps, and makes predictions through well-designed decoders and query embeddings.
For instance, L$^2$TKG~\cite{zhang2023learning} obtains entity and predicate embeddings via historical graph structure learning and latent predicates learning.
These embedding-based models are incapable of tackling auxiliary elements within n-tuples, thereby constraining their effectiveness in N-TKG reasoning. Moreover, their limited interpretability undermines the trust that users can place in them. 

In contrast to the embedding-based methods, multi-hop reasoning over TKGs can provide explainable predicted results, as well as achieve promising performance~\cite{sun2021timetraveler,zheng2023dream}. 
Among them, TLogic~\cite{liu2022tlogic} and TR-Rules~\cite{li2023tr} both propose to automatically mine temporal logical rules by performing temporal random walks over history. Building on this, TILP~\cite{xiong2024tilp} further integrates temporal random walks with temporal feature modeling, enabling more efficient learning of temporal logical rules. 
On the other hand, some works try to use the RL framework to search over historical TKGs and find evidential paths. 
Cluster~\cite{li2021search} first identifies potential temporal paths from historical facts through RL, and subsequently employs a GCN to deduce answers.
TITer~\cite{sun2021timetraveler}, which also adopts a RL framework, designs a relative time encoding function and time-shaped reward to automatically mine temporal patterns.
More recently, DREAM~\cite{zheng2023dream} leverages the generative adversarial imitation learning mechanism to minimize the dependence on handcrafted reward functions.
However, these approaches are designed for quadruple-based TKGs rather than N-TKGs, and thus cannot be directly applied to n-tuples and have to take adaptation measures.

Regarding the N-TKG reasoning task, the relevant method is relatively scarce. NE-Net~\cite{hou2023temporal} performs prediction based on evolutional representations of entities and predicates, which are learned through modeling both the complete information and the core information within historical facts. Although NE-Net shows great power in making predictions, its black-box characteristic results in poor interpretability and untrustable performance.

\section{Preliminaries}
An N-TKG $\mathcal{G}$ can be represented as a sequence of KG snapshots, i.e., $\mathcal{G}=\{\mathcal{G}_1,\mathcal{G}_2,...,\mathcal{G}_n,...\}$.
The KG at timestamp $t$ can be formalized as $\mathcal{G}_t=\{{V}_{ent},{V}_{pred},V_{role},{F}_t\}$, where ${V}_{ent}$, ${V}_{pred}$,  ${V}_{role}$,and ${F}_t$ denote finite sets of entities, relations, roles, and facts occurring at timestamp $t$, respectively.  
Each fact $f\in {F}_t$ is of the form $r(\rho_1\!:\!e_1,\rho_2\!:\!e_2,...,\rho_n\!:\!e_n,t)$, in which $r\in {V}_{pred}$ is the predicate; each $e_i \in {V_{ent}}$ represent the entity involved in this fact and plays as the role, as $\rho \in V_{role}$; 
$n$ stands for the number of entities participating in $f$, and $t$ represents the timestamp of the n-tuple. 
Particularly, in each fact, $e_1$ and $e_2$ are taken as core entities, while others are the auxiliary entities. 
The core elements of a fact encompass not only these two core entities but also the predicate, which specifies the relationship between them and defines the type of the fact.

Based on the historical KG sequence $\mathcal{G}_{< t}=\{\mathcal{G}_1,\mathcal{G}_2,...,\mathcal{G}_{t-1}\}$, the N-TKG reasoning task aims to predict the missing entity given a query in the form of $f_q\!=\!r_q(\rho_1\!:\!e_q,\rho_2\!:?,...,\rho_n\!:\!e_n,t)$. In this paper, we only focus on predicting the core entities within future facts.

\section{Methodology}
MT-Path formulates the multi-hop reasoning process as a sequential decision problem and solves it utilizing the RL framework. 
As illustrated in Figure~\ref{fig:model}, the MT-Path consists of two parts, the N-TKG environment and the agent. 
For each query at timestamp $t$, the agent first leverages a semantic component (``SemComp'' in Figure~\ref{fig:model}) to model the semantic dependency among historical n-tuples via an auxiliary element-aware GCN. 
Subsequently, starting from the query core entity, the agent selects an outgoing action based on the mixture policy-driven action selector (``ActSelector'' in Figure~\ref{fig:model}). 
The agent traverses to a new entity until it arrives at the target entity or reaches maximum path length $L$.
\subsection{The Framework of Reinforcement Learning}
Searching over N-TKGs is challenging due to the complexity of n-tuples. 
Typically, an entity may engage in numerous historical facts, leading to a vast and unwieldy action space. However, not all of these facts contribute equally to forming meaningful reasoning paths. Instead, it is the fact the entity serves as a core element that is crucial. To this end, we impose restrictions on the search process within the N-TKG environment. 
Specifically, the process of searching over the N-TKG environment can be viewed as a Markov Decision Process (MDP), which has the following components.

{\bf States.}  Each state is a tuple $s=(e_l,\tau_l, f_q)\in {S}$, where $(e_l, \tau_l)$ represents the entity, and the timestamp at the current reasoning step $l$, and $f_q$ denotes the query fact.
Note that $(e_l, \tau_l)$ can be viewed as state-dependent information while $f_q$ is the global context shared by all states.
The agent starts from the core entity of the query fact, so the initial
state is $s_0=(e_q, t, f_q)$.

{\bf Actions.} The set of candidate actions ${A}_l\in {A}$ at step $l$ consists of valid actions of the entity $e_l$ in $\mathcal{G}$. Considering that only those facts in which the entity $e_l$ serves as a core element are crucial, we restrict valid actions of $e_l$ to the historical facts wherein $e_l$ acts as the core entity. In this way, only meaningful actions are considered.
Concretely,\par
{\small{}{}{}{} 
\begin{equation}
\setlength{\abovedisplayskip}{-5pt} 
\setlength{\belowdisplayskip}{3pt}
    \begin{split}
    {A}_l=\big\{f&=r(\rho_1\!:\!e_l,\rho_2\!:\!e_2,...,\rho_n\!:\!e_n,\!\tau)\in \mathcal{G}\\
    &|\,\tau\,\le \,\tau_l\,,\,\tau\,<\,t\big\}.
    \end{split}
    \vspace{-2mm}
\end{equation}
}{\small\par}
\noindent To give the agent the option to terminate, a self-loop fact is added to every ${A}_l$.

{\bf Transition.} The transition function $\delta:{S}\times {A}\rightarrow {S}$ updates the state $s_l$ to a new state $s_{l+1}=(e_{l+1},\tau_{l+1},f_q)$ based on the action selected by the agent, i.e., $a_{l+1}$. Here, $e_{l+1}$ is the core entity in $a_{l+1}$ that differs from the entity $e_l$ within the previous state $s_l$.

{\bf Reward.}
The agent receives a terminal reward of 1 if it arrives at a correct target entity at the end of the search, and 0 otherwise. 
Following~\citet{sun2021timetraveler}, we additionally introduce a time-shaped reward, which can help the agent understand the distribution of correct entities over time.
\par
{\small{}{}{}{} 
\begin{equation}
\setlength{\abovedisplayskip}{-5pt} 
\setlength{\belowdisplayskip}{3pt}
    R(e_L)=(1+p_{\Delta t_L})\mathbb{I}(e_L==e_2),
    \vspace{-1mm}
\end{equation}
}{\small\par}
\noindent where $e_L$ and $\tau_L$ are the entity and the timestamp at the $L$ step; $\Delta t_L=t-\tau_L$ denotes the time interval among $t$ and $t_L$;  $p_{\Delta t_L}$ is the probability of the answer entity occurring at timestamp $\tau_L$ which can be  estimated by the Dirichlet distribution. $\mathbb{I}(P)$ is the indicator function that obtains 1 if $P$ is true, otherwise 0.
\subsection{The Semantic Component} 
When determining where to go, simply focus on the information within the query, and each candidate action may be insufficient. Actually, facts can influence each other via shared entities, which exhibit structural dependencies within N-TKGs. Such kinds of dependencies contain global information of a fact, which is important for an agent to make precise predictions~\cite{zheng2023dream}.
Motivated by this, MT-Path employs an auxiliary element-aware GCN to capture these semantic dependencies among historical facts. 

Since the number of historical facts is huge, we only consider n-tuple facts with timestamps from $t-m$ to $t-1$, and transform them into a background knowledge graph: 
\par
{\small{}{}{}{} 
\begin{equation}
\setlength{\abovedisplayskip}{-5pt} 
\setlength{\belowdisplayskip}{3pt}
\begin{split}
\mathcal{G}_{bg}^{t}=\bigcup_{\tau=t-m}^{t-1}&\big\{\hat{f}=r(\rho_1\!:\!e_1,\,\rho_2\!:\!e_2,..,\,\rho_n\!:\!e_n)\,|\\
&\exists\,r(\rho_1\!:\!e_1,\rho_2\!:\!e_2,..,\rho_n\!:\!e_n,\tau)\in {F}_\tau\big\}.
\end{split}
\vspace{-2mm}
\end{equation}
}{\small\par}

Based on the constructed background graph, this component employs a $w$-layer auxiliary element-aware GCN~\cite{galkin2020message} to capture the semantic dependencies among historical n-tuples. 
Specifically, at timestamp $t$, the core entity $u$ at layer $k\in [0,w-1]$ gets information from the other core entity under a message-passing framework with the augmented predicate embedding $\hat{\boldsymbol{r}}^{t}$ at layer $k$, and obtains its embedding at the next $k+1$ layers, i.e.,\par
{\small{}{}{}{} 
\begin{equation}
\setlength{\abovedisplayskip}{-5pt} 
\setlength{\belowdisplayskip}{3pt}
    \boldsymbol{h}^{t,k}_u=
    \psi\Big(\sum_{\hat{f}_{u,v}\in \mathcal{G}_{bg}^{t}}\!\mathbf{W}^k_{0}(\hat{\boldsymbol{r}}^{t}+\boldsymbol{h}_v^{t,k-1})+\mathbf{W}_1^{k-1} \boldsymbol{h}_u^{t,k}\Big),
    \vspace{-1mm}
\end{equation}
}{\small\par}
\noindent where $\hat{f}_{u,v}$ represents a fact that takes $u$ and $v$ as the core entities and takes $r$ as the predicate; $\boldsymbol{h}_u^{k,t}$ and 
$\boldsymbol{h}_v^{k,t}$ denotes the embeddings of the entity $u$ and $v$ obtained after $k$ layers, respectively;
$\mathbf{W}_{0}^{k}$ and $\mathbf{W}_{1}^{k}$ are the parameters in the $k$-th layer; $\psi(\cdot)$ is an activation function.
The augmented predicate embedding $\hat{\boldsymbol{r}}^{t}$ is defined as the weighted sum of the predicate embedding $\boldsymbol{r}$ and the auxiliary embedding $\boldsymbol{h}_{aux}^{t}$:
\par
{\small{}{}{}{} 
\begin{equation}
\setlength{\abovedisplayskip}{-5pt} 
\setlength{\belowdisplayskip}{3pt}
\hat{\boldsymbol{r}}^{t}=w*\boldsymbol{r}+(1-w)*\boldsymbol{h}_{aux}^{t},
\vspace{-1mm}
\end{equation}
}{\small\par}
\noindent where $w\in (0,1)$ is the weight hyperparameter; $\boldsymbol{h}_{aux}^t$ represents the information of all auxiliary role-entity pairs within the fact $\hat{f}_{u,v}$, i.e., $aux=\{(\rho_i\!:\!e_i)\}_{i=3}^n$, and can be calculated through a one-layer GCN: \par
{\small{}{}{}{} 
\begin{equation}
\setlength{\abovedisplayskip}{-5pt} 
\setlength{\belowdisplayskip}{3pt}
\boldsymbol{h}_{aux}^{t}=\psi\bigg(\sum_
    {(\rho,e)\in {aux}}
    \mathbf{W}_3(\boldsymbol{\rho}+\boldsymbol{h}_{e}^t)\bigg).
    \vspace{-1mm}
\label{eq:aux}
\end{equation}
}{\small\par}
\noindent Note that, for the first layer, both the entity embeddings  $\mathbf{H}$ and the predicate embeddings $\mathbf{R}$ are randomly initialized. After the $w$-layers message passing, all updated entity embeddings $\mathbf{H}_{t}$ are used as the input of the following action selector.

\subsection{The Mixture Policy-Driven Action Selector}
As aforesaid, besides the time information, there are diverse kinds of semantic information within n-tuples, i.e., the entity-irrelevant information within the predicate, the information about core elements, and the complete information about the entire n-tuples, all of which are crucial for making precise predictions.
Thus, the main design principle of MT-Path is to determine the next step by synthesizing both the semantic and the time information within the query, the candidate actions, as well as existing reasoning paths.
Specifically, MT-Path utilizes a mixture policy-driven action selector to determine the next hop. At each step, the selector decomposes the action selection strategy into three low-level policies with similar structures, i.e., the predicate-focused policy (``P-Policy'' in Figure~\ref{fig:model}), the core-element-focused policy (``C-Policy'' in Figure~\ref{fig:model}), and the whole-fact-focused policy (``F-Policy'' in Figure~\ref{fig:model}), which calculate each candidate' probability based on the information within the predicates, the core elements, and the entire facts, respectively, as well as the time information.
After that, the outputs of these three low-policies are dynamically integrated so as to obtain the final probability of each candidate action. In the following equations, we denote $P,C,F$ to represent the P-Policy, the C-Policy, and the F-Policy, respectively.

Typically, the more recent candidate actions carry greater significance than those from the distant past. As a result, MT-Path models the time information within the action $a_i$ via encoding the time interval between its timestamp $\tau_i$ and the query time $t$, i.e.,
    $\mathbf{\Phi}(\Delta t_i)=cos((t-\tau_i)\boldsymbol{w}_t+\boldsymbol{b}_t)$.
Here $\boldsymbol{w}_t$ and $\boldsymbol{b}_t$ are both learnable parameters.
In each low-level policy, the embedding of the action $a_i$ is the concatenation of the embedding of the elements that the policy focuses on and the corresponding time interval embedding. 
More details, in the P-Policy, the action embedding of action $a_i$ is the concatenation of the predicate embedding and the time interval embedding, i.e., $\boldsymbol{a}^P_i=[\boldsymbol{r}_{i};\mathbf{\Phi}(\Delta t_i)]$.
Compared to $\boldsymbol{a}^P_i$, the C-Policy additionally extends the action embedding by including the core entity embedding $\boldsymbol{h}_{e_i}^t$ at timestamp $t$, i.e., $\boldsymbol{a}_i^C = [\boldsymbol{r}_i; \boldsymbol{h}_{e_i}^t; \mathbf{\Phi}(\Delta t_i)]$. Note that $e_i$ is the core entity differing from the entity in the previous state. 
Based on $\boldsymbol{a}_i^C$, the F-Policy further incorporates the information within auxiliary role-entity pairs of $a_i$, i.e., $aux_i$, into the action embedding, i.e., $\boldsymbol{a}_i^F = [\boldsymbol{r}_i; \boldsymbol{h}_{e_i}^t; \boldsymbol{h}_{aux_i}^t; \mathbf{\Phi}(\Delta t_i)]$. Here, $\boldsymbol{h}_{aux_i}^t$ is derived using Equation~\ref{eq:aux}.

Formally, the search history $h_l=(a_0,a_1,...,a_{l-1})$ consists of the sequence of visited facts. 
Each low-level policy $I$, where $I\in \{P, C, F\}$,  utilizes an LSTM (``$I$-His Encoder'' in Figure~\ref{fig:model}) to encode the relevant information in the search history $h_l$ into a hidden vector $\boldsymbol{h}_l^I$: \par
{\small{}{}{}{} 
\begin{equation}
\setlength{\abovedisplayskip}{-5pt} 
\setlength{\belowdisplayskip}{3pt}
\begin{split}
    &\boldsymbol{h}_0^I=LSTM(\mathbf{0},\boldsymbol{a}_0^I),\\
    &\boldsymbol{h}_l^I=LSTM(\boldsymbol{h}_{l-1}^I,\boldsymbol{a}_{l-1}^I]).
    \end{split}
    \vspace{-1mm}
\end{equation}
}{\small\par}
\noindent Subsequently, a two-layer feed-forward network is adopted to calculate the probability distribution over all possible actions in ${A}^I_l$:\par
{\small{}{}{}{} 
\begin{equation}
\setlength{\abovedisplayskip}{-5pt} 
\setlength{\belowdisplayskip}{3pt}
\pi_\theta^I(a_l|s_l)\!=\!\sigma(\mathbf{A}^{I}_l\!\times\!\mathbf{W}^{I}_2 ReLU(\mathbf{W}^{I}_1[\boldsymbol{h}^I_l;\boldsymbol{info}^I_q])),
\end{equation}
}{\small\par}
\noindent where $\mathbf{A}^I_l$ is the stack of all actions embeddings $\mathbf{a}_i^{I}$ in $A_l$ and $\sigma$ denotes the softmax operator; $\mathbf{W}^{I}_1$ and $\mathbf{W}^{I}_2$ are both learnable parameters; $\boldsymbol{info}^I_q$ represents the concatenated embeddings of elements within the query that the low-level policy $I$ focuses. 
Specifically, for $I\in \{P,C,F\}$, $\boldsymbol{info}_q^I$ equals to $[\boldsymbol{r}_q]$, $[\boldsymbol{r}_q;\boldsymbol{h}^t_{e_q}]$, and $[\boldsymbol{r}_q;\boldsymbol{h}^t_{e_q};\boldsymbol{h}_{aux_q}^t]$, respectively. Here, $aux_q$ denotes all auxiliary role-entity pairs in the query $f_q$, and its embedding $\boldsymbol{h}_{aux_q}^t$ is calculated by Equation~\ref{eq:aux}.

To integrate the outcomes calculated by the above three low-level policies, i.e., $\pi_\theta^P$, $\pi_\theta^C$ and $\pi_\theta^F$, MT-Path utilizes an MLP-based gate to learn the importance weights of the three kinds of information, and adaptively combine them in order to obtain the final probability:\par
{\small{}{}{}{} 
\begin{equation}
\setlength{\abovedisplayskip}{-5pt} 
\setlength{\belowdisplayskip}{3pt}
\mathbf{W}_{gate}=\sigma(\mathbf{W}_g[\boldsymbol{h}_l^P;\boldsymbol{h}_l^C;\boldsymbol{h}_l^F;\boldsymbol{r}_q;\boldsymbol{h}_{e_q}^t;\boldsymbol{h}^t_{aux_q}]).
\end{equation}
}{\small\par}
\noindent Finally, the final probability of action $a_i$ is defined as below:\par
{\small{}{}{}{} 
\begin{equation}
\setlength{\abovedisplayskip}{-5pt} 
\setlength{\belowdisplayskip}{3pt}
\pi(a_i|s_l)=\sigma(\mathbf{W}_{gate}[\pi^P_\theta;\pi^C_{\theta};\pi^F_\theta]).
\vspace{-1mm}
\end{equation}
}{\small\par}
\begin{table}[t]
\centering
\resizebox{0.9\linewidth}{!}{
\begin{tabular}{ccccccccccccccc}
\toprule
{Dataset}&{$|R|$}&{$|E|$} &{\#Train}&{\#Valid}&{\#Test}&{Time Interval}\\
\midrule
NWIKI  &22 &17,481  &108,397 &14,370 &15,591&1 year \\
NICE  &20 &10,860   &368,868 &5,268 &46,159&24 hours \\
\bottomrule
\end{tabular}
}
\caption{The statistics of the datasets.}
\label{tab:datasets}
\vspace{-5mm}
\end{table}

\subsection{Optimization and Training.}
The parameters of MT-Path are trained by maximizing the expected reward over all queries in the training set,
\par
{\small{}{}{}{} 
\begin{equation}
\setlength{\abovedisplayskip}{-5pt} 
\setlength{\belowdisplayskip}{3pt}
\begin{split}
    J(\theta)=&\mathbb{E}_{f\in\mathcal{G}}[\mathbb{E}_{a_0,...,a_T\sim \pi_{\theta}}[R(e_L\\
    &|r,\rho_1,e_1,\rho_2,...,\rho_n,e_n,t,\mathcal{G}_{1:t})]].
    \end{split}
\vspace{-1mm}
\end{equation}
}{\small\par}
\noindent The optimization is then performed by using REINFORCE~\cite{williams1992simple} algorithm.

\begin{table*}[t]
\centering
\resizebox{0.6\linewidth}{!}{
\begin{tabular}{cc|cccc|ccccc}
\toprule
\multicolumn{2}{c|}{\multirow{2}*{Model} }& \multicolumn{4}{c|}{NWIKI} &\multicolumn{4}{c}{NICE}\\ 
 \cmidrule{3-10}
 \multicolumn{2}{c|}{}&  H@1  &   H@3 &  H@10  &  MRR  &   H@1 &  H@3  &  H@10  &   MRR\\
\hline
\multirow{6}*{\rotatebox{0}{C1}}&NALP &10.59 &10.59&22.52 &14.86
&14.66 &26.17&43.40 &23.96\\
&NeuInfer&19.83&25.22 &28.94 &23.08
&13.77 &28.35 &47.56 &24.78\\
&HINGE &19.10 &23.47 &25.91 &21.74
&2.92 &21.04 &42.83 &16.01 \\
&RAM&31.42 &33.36 &34.36&32.63
&8.37 & 16.41 &27.13 &14.38 \\
&HypE &24.91 &25.39 &25.75 &25.26
&19.16 &37.22 &56.33&31.50\\
&Hy-Transformer&33.40 &35.85 &37.84 &34.97
&28.51 &44.49 &61.11 &39.47 \\
\midrule
\multirow{6}*{\rotatebox{0}{C2}}
&RENET &33.56 &38.41 &41.28 &36.57 
&33.43 &47.77 &63.06 &43.32 \\
&CyNet 
&44.12 &64.71 &67.65 &53.12 
&26.61 &41.63 &56.22 &36.81 \\
&RE-GCN 
&46.25 &65.13 &72.31 &56.78 
&37.33 &53.85 &68.27 &48.03\\
&GHT 
&30.71 &37.78 &39.94 &34.57 
&26.61 &41.63 &56.22 &36.81 \\
&CEN &30.28 &45.20 &61.04 &40.61 
&33.32 &49.29 &64.65 &43.98 \\
&TiRGN 
&50.61 &68.24 &81.13 &61.10 
&34.82 &51.54 &66.47 &45.66\\
\midrule
\multirow{6}*{\rotatebox{0}{C3}}
&TITer&64.68	&81.75	&83.04	&73.30
&37.79	&53.23	&66.32	&47.74 \\
&Cluster &41.10	&58.85	&\bf 95.59	&55.55 
&31.62	&48.01	&65.80	&42.90\\
&xERTE &39.64	&51.91	&58.71	&46.46
&33.75	&49.29	&62.11	&43.20\\
&TLogic&70.67	&81.95	&82.66	&76.36
&33.55	&48.27	&61.22	&43.04 \\
&LCGE &33.90 &37.09 &40.14 &36.18
&33.53 &47.28 &62.14 &43.16\\
&TR-Rules &58.20 &64.63   &71.53 &62.57
&30.82 &44.56 &56.64 &39.56 \\
\midrule
\multirow{1}*{\rotatebox{0}{C4}}
&NE-Net  &{66.87} &{76.08}
&{80.29} &{72.03} 
&{38.36} &{54.18} 
&{\bf 69.99} &{48.98} \\
\midrule
&MT-Path &\bf 78.98 &\bf 81.90 &83.43 &\bf 80.69
&\bf 40.16	&\bf 55.57	&67.99	&\bf 49.91 \\
\bottomrule
\end{tabular}
}
\caption{Experimental results on N-TKG reasoning compared with static n-tuple reasoning models (C1), embedding-based TKG reasoning models (C2), multi-hop-based TKG reasoning models (C3), and embedding-based N-TKG reasoning models (C4). }
\vspace{-5mm}
\label{table:mainpred}
\end{table*}

\section{Experiments}
\subsection{Experimental Setup}
{\bf Datasets.} To evaluate the effectiveness of MT-Path, we conduct experiments on two N-TKG datasets, i.e., NICE and NWIKI~\cite{hou2023temporal}. 
The NICE dataset is derived from the large-scale event-based database, ICEWS, from Jan 1, 2005 to Dec 31, 2014. 
The NWIKI dataset, derived from Wikidata~\cite{vrandevcic2014wikidata}, is a knowledge base with a time granularity of years.
The statistics of these datasets are presented in Table~\ref{tab:datasets}.

{\bf Evaluation Metrics.}
We adopt two widely used metrics, i.e., standard Mean Reciprocal Rank (MRR) and Hits@$\{ 1, 3, 10 \}$, to evaluate the reasoning performance.
Following~\citet{hou2023temporal}, we perform time-aware filtering where only the facts occurring at the same timestamp as the query are filtered from the ranking list of corrupted facts.

\begin{table}[t]
\centering
\resizebox{0.85\linewidth}{!}{
\begin{tabular}{c|c|c|c|c|c|ccccccccc}
\toprule
\multirow{3}{*}{Variants}&\multicolumn{3}{c}{NWIKI} &\multicolumn{3}{|c}{NICE}\\
\cmidrule{2-4}\cmidrule{5-7}
&H@1 &H@3&MRR&H@1 &H@3&MRR\\
\midrule

-SC  &74.48 &\textbf{82.11} &78.52 &35.56 &49.34 &{45.05} \\
-CP  &78.38 &82.06 &80.46 &39.16 &53.36 &{48.46} \\
-PP  &78.83 &81.88 &80.64 &38.53 &52.83 &{47.88} \\
-FP  &61.20 &80.90 &71.19 &32.23 &47.65 &{42.58} \\
-GA  &74.42 &82.11 &78.52 &34.03 &48.06 &{43.67} \\
MT-Path  &\textbf{78.98} &{81.90} &\textbf{80.69} &\textbf{40.16} &\textbf{55.57}&\textbf{49.91} \\
\bottomrule
\end{tabular}
}
\caption{Ablation studies of the proposed MT-Path.}
\label{tab:ablation}
\vspace{-5mm}
\end{table}

{\bf Baselines.}
We compare MT-Path with the following four kinds of models: static n-tuple reasoning models, embedding-based TKG reasoning models, multi-hop-based TKG reasoning models, and embedding-based N-TKG reasoning models. In the first category, NALP~\cite{guan2019link}, NeuInfer~\cite{guan2020neuinfer}, HINGE~\cite{rosso2020beyond}, RAM~\cite{liu2021role}, HypE~\cite{fatemi2021knowledge} and Hy-Transformer~\cite{yu2021improving} are compared.
In the second category,  RENET~\cite{jin2020Renet}, CyNet~\cite{zhu2020learning}, RE-GCN~\cite{li2021temporal}, GHT~\cite{sun2022graph}, CEN~\cite{li2022complex}, and TiRGN~\cite{li2022tirgn} are taken as baselines.
In the third category,  
we choose TITER~\cite{sun2021timetraveler}, Cluster~\cite{li2021search}, xERTE~\cite{han2020xerte}, TLogic~\cite{liu2022tlogic}, LCGE~\cite{niu2023logic}, and TR-Rules~\cite{li2023tr} as our baselines. 
In the last category, NE-Net~\cite{hou2023temporal} is chosen as our baseline.

{\bf Implementation Details.} 
For both datasets, the dimensions of predicate embeddings, entity embeddings, and time interval embeddings are set to 100, 80, and 20, respectively. The maximum reasoning step $L$ is set as 3. The number of LSTM layers is set to 2 and the output dimension of the LSTM unit is set to 100. 
For SemComp, the latest KG number $m$ is set to 5 and 1 for NICE and NWIKI, respectively. 

\begin{table*}[t]
\centering
\resizebox{0.7\linewidth}{!}{
\begin{tabular}{c|c|c}
\toprule
\multicolumn{1}{c}{Query} & \multicolumn{1}{c}{Reasoning Path} & \multicolumn{1}{c}{Answer} \\
\midrule 
\makecell[l]{\textbf{Yield}\\ 
(Yielder: $E_1$, Demander: ?,\\
Way: $E_2$, Place: $E_3$, $t$)}
& \makecell[l]{\textbf{Coerce}\\
(Coercer: $E_1$, Coercee: $E_4$, Way: $E_5$, Place: $E_3$, $t-1$)\\
\textbf{Disapprove}$^{-1}$\\
(Disapprover: $E_4$, Target: $E_1$, Way: $E_6$, Place: $E_3$, $t-2$)\\
\textbf{Disapprove}\\
(Disapprover: $E_1$, Target: $E_4$, Way: $E_6$, Place: $E_3$, $t-3$)}
& $E_4$ \\
\midrule
\makecell[l]{\textbf{Reject}\\
(Rejector: $\bm{E_1}$, Rejectee: ?,\\
Content: $E_2$, $t$)}
& \makecell[l]{\textbf{Disapprove}$^{-1}$\\
(Disapprover: $E_3$, Target: $\bm{E_1}$, Way: $E_4$, $t-1$)\\
\textbf{Reject}\\
(Rejector: $E_3$, Rejectee: $E_5$, Content: $E_2$, $t-2$)}
& $E_5$ \\
\midrule
\makecell[l]{\textbf{Engage in Diplomatic Coop}\\ 
(Cooper1: $E_1$, Cooper2: $?$,\\
Way: $E_2$, Place: $E_3$, $t$)}
& \makecell[l]{\textbf{Express Intent to Coop}\\
(Volunteer: $E_1$, Target: $E_4$, Way: $E_6$, Place: $E_5$, $t-1$)}
& $E_4$ \\
\bottomrule
\end{tabular}
}
\caption{Case Study of MT-Path. In the second case, $\bm{E_1}$ is an unseen entity.}
\label{tab:case}
\vspace{-5mm}
\end{table*}
\subsection{Experimental Results}
The experimental results of MT-Path and all baselines on N-TKG reasoning are presented in Table~\ref{table:mainpred}. 
It can be seen that MT-Path outperforms all baselines on two datasets in terms of MRR, Hits@1, and Hits@3, which verifies the superiority of MT-Path. 
Specifically, MT-Path outperforms all static n-tuple reasoning methods, because it can capture the time information within each n-tuple and utilizes such kind of information to conduct reasoning.
Further, it can be observed that the performance of MT-Path is much higher than the embedding-based TKG reasoning methods and the multi-hop-based TKG reasoning methods. This phenomenon occurs as MT-Path employs the F-Policy to capture the complete information about the entire n-tuples, and additionally utilizes such kind of information to conduct reasoning. 
Compared with the most related baseline, i.e., NE-Net, MT-Path still shows better performance. 
However, it can be noticed that MT-Path usually performs worse in terms of Hits@10, compared with  NE-Net and Cluster. 
This is because MT-Path relies heavily on strict searching within historical n-tuples to find evidential paths that lead to the target entities. 
In contrast, both NE-Net and the final stage of Cluster map all entities and predicates into a unified embedding space to capture inner connections.
This relaxes the strict searching restriction and thus both NE-Net and Cluster can achieve a higher recall level like Hits@10.

\subsection{Ablation Study}
To further analyze how each part of MT-Path contributes to the final results, we conduct ablation studies on the NICE dataset. The results are summarized in Table~\ref{tab:ablation}.

To verify the effectiveness of semantic component (denoted as -SC), we simply utilize randomized entity embeddings as the input of the action selector. It can be observed that -SC results in worse performance compared with MA-Path, which illustrates the necessity of additionally capturing the semantic dependencies among historical facts.  

To further analyze the importance of the mixture policy, we deactivate three low-level policies, i.e., the P-Policy, the C-Policy, and the F-Policy, denoted as -PP, -CP, and -FP, respectively.
It can be observed that MT-Path shows better performance than these three variants. Also, we notice that -FP generates a bigger performance drop compared with -PP and -CP. This is because F-Policy additionally captures the auxiliary information within the entire n-tuples, which contain supplementary descriptive details of a fact.

Additionally, to verify the necessity of the adaptive aggregation unit (-GA in Table~\ref{tab:ablation}), we simply treat each low-level policy as having equal importance. The underperformance of -GA demonstrates that adaptively integrating the outcomes generated by the low-level policies can help determine the next hop better.

\begin{figure}
  \centering  
   \includegraphics[width=0.85\linewidth]{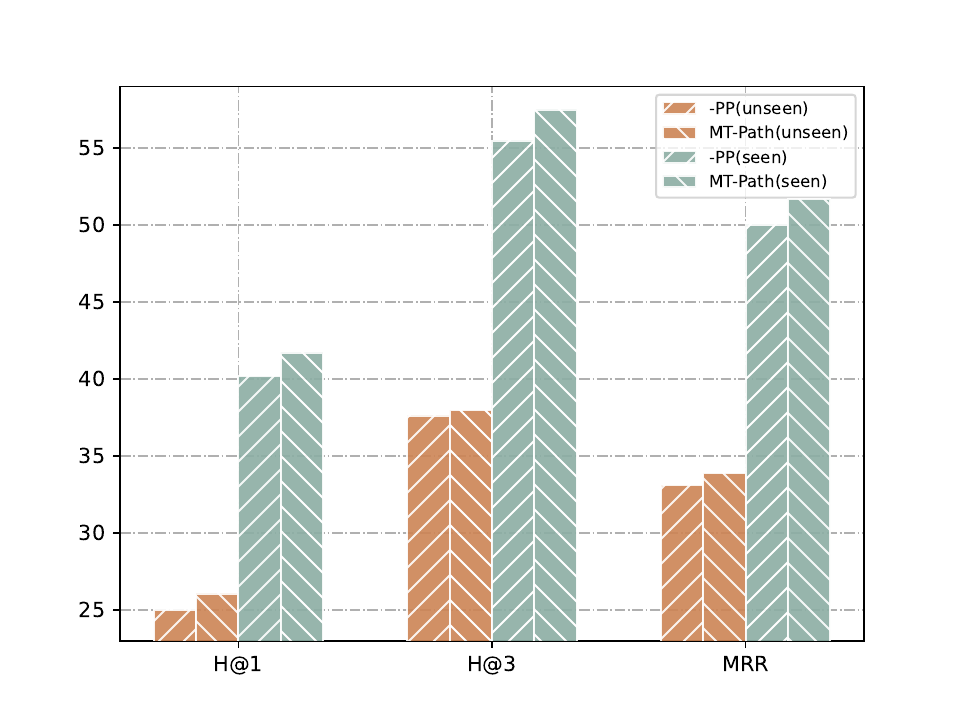}
   \caption{Performance of MT-Path over queries involving seen and unseen entities.}
   \label{fig:unseen}
    \vspace{-5mm}
\end{figure}

\subsection{Detailed Analysis}
To analyze the effectiveness of MT-Path in the inductive setting, we split queries in NICE into two categories: those involving seen and unseen entities. 
As shown in Figure~\ref{fig:unseen}, MT-Path outperforms its variant without the P-Policy (``-PP" in Figure~\ref{fig:unseen}) when dealing with facts involving unseen entities. 
This is because, without the P-Policy, the model relies heavily on the embeddings of these unseen entities, which have not been adequately optimized during the training stage. Conversely, the P-Policy in MT-Path emphasizes the entity-irrelevant information within the predicate, thereby alleviating the negative effects of unseen entities.

\subsection{Case study}
In Table~\ref{tab:case}, we provide three cases in which MT-Path correctly predicts the answer entity. 
For each query, the reasoning path with the highest score by MT-Path is provided.
From the first case, we can observe that MT-Path takes actions that take place at the same location as the query.
This phenomenon underscores MT-Path's capability to effectively harness the auxiliary information within candidate actions, particularly when it aligns with the information provided in the query. 
In the second case, the core entity $E_1$ in the query is a new emerging entity. Despite this, MT-Path manages to make an accurate prediction and successfully identifies a plausible reasoning path from the history. This can be attributed to the P-Policy, as it mainly focuses on the entity-irrelevant information within the predicate and can help mitigate the influence of the unseen entity.
From the last case, we can see that the auxiliary entities in the query ($E_2$ and $E_3$) differ from those in the action ($E_5$ and $E_6$).
This suggests that MT-Path can determine the next hop based on not only the whole information in both query and historical facts, but also the information within the core elements and the predicate. Such ability can help MT-Path tackle complex prediction scenarios.
Furthermore, from all these cases, it can be observed that MT-Path can mine reasoning paths with different lengthy to support its predictions.

\section{Conclusions}
In this paper, we introduced MT-Path, a new RL-based method to traverse the historical n-tuples and conduct multi-hop reasoning over N-TKGs.
MT-Path decomposes the action selection strategy into three low-level ones, to determine the next hop based on the entity-irrelevant information within the predicates, the information about the core elements, and the complete information about the entire n-tuples. It further employs an auxiliary element-aware GCN to capture the semantic dependencies among historical n-tuples. Experiments on two benchmarks demonstrate the advantages of MT-Path on N-TKG reasoning. Moreover, the explicit paths found by MT-Path further provide interpretability for the reasoning results.

\section*{Limitations}
The limitations of this work can be concluded into two points: (1) MT-Path does not explore the prediction of auxiliary entities. Designing a model capable of simultaneously predicting both core and non-core entities is a good direction for future studies.
(2) MT-Path incorporates multiple components to find evidential paths, which may affect its scalability in real-time scenarios.
\bibliography{acl}

\appendix



\end{document}